\documentclass[conference]{IEEEtran}
\IEEEoverridecommandlockouts
\usepackage{cite}
\usepackage{amsmath,amssymb,amsfonts}
\usepackage{amssymb}
\usepackage{algorithmic}
\usepackage{graphicx}
\usepackage{textcomp}
\usepackage{xcolor}
\newcommand{\invamalg}{\mathbin{\text{\rotatebox[origin=c]{180}{$\amalg$}}}}
\def\BibTeX{{\rm B\kern-.05em{\sc i\kern-.025em b}\kern-.08em
    T\kern-.1667em\lower.7ex\hbox{E}\kern-.125emX}}
\begin{document}

\title{A Roadmap to Domain Knowledge Integration in Machine Learning\\

}

\author{

\IEEEauthorblockN{Himel Das Gupta}
\IEEEauthorblockA{\textit{Department of Computer Science} \\
\textit{Texas Tech University, Lubbock, TX} \\
Himel.Das@ttu.edu}

\and

\IEEEauthorblockN{Victor S. Sheng}
\IEEEauthorblockA{\textit{Department of Computer Science} \\
\textit{Texas Tech University, Lubbock, TX} \\
victor.sheng@ttu.edu}
}

\maketitle

\begin{abstract}
Many machine learning algorithms have been developed in recent years to enhance the performance of a model in different aspects of artificial intelligence. But the problem persists due to inadequate data and resources. Integrating knowledge in a machine learning model can help to overcome these obstacles up to a certain degree. Incorporating knowledge is a complex task though because of various forms of knowledge representation. In this paper, we will give a brief overview of these different forms of knowledge integration and their performance in certain machine learning tasks. 

\end{abstract}


\section{Introduction}
Deep learning is flourishing in the field of Artificial Intelligence (AI) for a long time. It is continuously wide-spreading its enormous success to the various fields of computer vision, autonomous vehicle driving, game playing, etc. Significant improvements have been made over the years to make Deep Leaning more robust.
Despite being robust, the performance of deep learning still highly depends on the input data set.  But even in this era of big data, it is highly intensive to feed huge datasets into a deep neural network. High reliability in data-driven learning may lead to inexplicable results. Integrating prior knowledge into a neural network can bring a more prominent solution in this regard.

We can find a famous debate in intellectual history. That is, how much is the human mind built-in? How much is the human brain formulated by experience? In \cite{b1} authors have posted their argument that artificial intelligence needs greater attention to innateness. The relation that connects machine learning with human knowledge is based on the process of learning. The idea of a neural network was indeed inspired by the architecture of the human brain. All these years, researchers have kept working on making the machine learning process as human as possible. In the context of machine learning, learning from the data without having any prior knowledge is one extreme point, whereas manually driven hard-coded knowledge also provides some help. The success rate of a deep learning model highly depends on sweet spots between these extremes. 

Though the idea of implementing knowledge inside a machine learning model is astonishing, often knowledge base suffers from a lack of completion. A novel method for knowledge graph completion has been proposed in \cite{b27} where the knowledge graph has been completed using pre-trained language models. In this paper, entities, relations, and triples have been considered as textual sequences making the knowledge graph a sequence classification problem. The idea of the framework  Knowledge Graph Bidirectional Encoder Representations from Transformer (KG-BERT) is to model these triples than can achieve state-of-art results in tasks like triple classification, relation prediction, and link prediction.   \\

Efficiency in natural languages processing tasks like entity recognition, sentiment analysis, and question answering can be improved with the help of representation of language pre-training \cite{b28}. But the problem persists as methods like this do not consider prior knowledge because of predicting the missing words through contexts only. Using masking strategies with the addition of e two
kinds of knowledge strategies: phrase-level strategy and entity-level strategy, authors have implemented a new model Enhanced Representation Through Knowledge Integration (ERNIE) \cite{b29}. In this model, a phrase or entity is considered one unit. In doing so, instead of masking only one word or one character, all the words in the same unit are masked during word representation training. Thus prior domain knowledge of entities is learned making the model more generalized and adaptive. Informative knowledge utilizing large-scale textual knowledge graphs can enhance further knowledge representation \cite{b30}.   

Depending on various types of data, different neural network architectures are used. The selection of such an architecture plays a vital role while training a machine learning model. We can see that Convolutional Neural Networks (CNN) work very well on data such as images and videos because this kind of data shows translational invariance. On the other hand, we can also see that Recurrent Neural Networks (RNN) work well on data with sequential structures. But in the end, it all comes down to the fact of how much data we have and how good they are. There can be several hindrances in real-world applications, such as \textbf{limited data}, \textbf{expensive data}, \textbf{poor quality data}, etc. The power of integrating domain knowledge comes in highlighting these scenarios. Such knowledge representation can be in the form of a physical model, constraints, scientific or physical laws, relations between features, etc.   

Deep learning as a statistical machine learning model has two primary ingredients, i.e., model architectures and learning algorithms. Domain knowledge can be integrated into either the model architectures or learning algorithms. Different ideas have been proposed over the years on how efficiently this knowledge integration can be done to improve the accuracy of a deep learning framework. In this paper, we pursue giving an overview of the current understanding of knowledge integration in a neural network. In addition, to arouse further research on this field, we focus on different opportunities and challenges of incorporating knowledge in deep learning. 

In a nutshell, we first describe the different representations of knowledge representation in section II, which will build the foundation of the importance of knowledge integration in machine learning. In sections III, IV, and V, we will focus on different ideas of formulating knowledge using mathematical rules in existing machine learning algorithms. In section VI, We have a discussion on opportunities and ideas for future research and development in this field. The paper will be ended with a conclusion in section VII.

\section{Nature of Knowledge Representation}
From a technical point of view, knowledge in computer science as well as in information technology can be expressed as the possession of information that can be quickly located. If we dive into the definition of AI, we can see that AI is actually the basis of human intelligence processes using algorithms that are built into a dynamic computing environment. The main motivation behind AI is to make computers and machines act like human beings. Knowledge in machine learning can be of various types, including entities (objects), events, performance, meta-knowledge, facts, and knowledge-base.

Domain knowledge in a machine learning framework can be classified into three groups: relational knowledge, logical knowledge, and scientific knowledge. 
Relational knowledge can be defined as a relation between entities. From a conceptual point of view, relational knowledge is binding between relation symbols and set of ordered tuples \cite{b2, b3}. For example, a relation-symbol can be something like \textit{father} which is bound to set of ordered pairs {(Alice, Bob), (Alice, Mary), ...}. This representation expresses the fact that Alice is the father of Bob and Mary. Relational knowledge like this can be exhibited by either a relational database or a knowledge graph. In a deep learning model, relational knowledge can be incorporated by a probabilistic graphical model to model relationships between entities. Probabilistic relational models and relational dependency networks can also be used as representations. Different work has been done in the embedding space of relational knowledge. Entities can be represented by a vector in a vector space, and then an entity can be predicted given its context. In \cite{b4} authors have shown that a significant enhancement in an information retrieval process can be achieved through word embedding.

Logical knowledge is another important class of knowledge. Logic can be expressed by propositional logic and first-order logic. Propositional logic can be used in artificial intelligence for planning, problem-solving, intelligent control, and most importantly for decision-making. It is all about Boolean functions and statements where there are more than just true and false values, including certainty as well as uncertainty and thus leading to the foundation for machine learning models. It is a useful tool for reasoning, but it has limitations because it cannot see inside prepositions, and cannot take advantage of relationships among them. An example of propositional logic can be like this, \textit{If you get more doubles than any other player then you will lose, or if you lose then you must have bought the most properties.} Logical knowledge can be encoded into probabilistic graphical models. Such an example can be found on Bayesian Network from Horn clauses, Probabilistic Context-Free Grammars\cite{b5}, Markov Logic Networks \cite{b6}.

Scientific knowledge is also an important part of knowledge that we can use as prior knowledge to a machine learning model. The general laws of physics, math, and astrology, all are part of scientific knowledge that we can bind into a neural network architecture. 

In the following sections, we further introduce the three groups of knowledge respectively in a little more detail.

\section{Relation Knowledge as Domain Knowledge Integration}
Tasks like reading comprehension and question-answering have been popular in natural language processing. There has been significant development of developing a deep neural network to extract answers in question-answering tasks. Most of these deep neural networks are specially developed to extract answers from only text or only from Knowledge Bases (KB). The problem persists when the KB is noisy or missing information. Unstructured text alone itself is also not prominent enough to draw a good conclusion. Integrating knowledge bases as entities to entity-linked text named \textbf{GRAFT-Net} is a novel idea to be successful in difficult question answer tasks \cite{b7}. 
Let's consider a graph, $\mathcal{G} = (\mathcal{V}, \mathcal{E})$ where vertices $\mathcal{V}$ represent entities and edges $\mathcal{E}$ represent the relations between these entities. The main objective is to learn a function $y_v = f(v) \forall v \in \mathcal{V}$ such that $y_v \in {0,1}$ and $y_v =1 $ if and only if $v$ is a answer set of $q$ where $q =(w_1, w_2,..., w_T)$ is a natural language question. The softmax representation of this relation is:

\begin{equation}
P(y_v=1 | \mathcal{G},q) = \frac{exp h_{q}^{T} h_v}{\sum_{v^{'}}exp h_{q}^{T}h_{v^{'}}}    
\end{equation}
where $h_q$ means the question representation from a long-short-term memory (LSTM) network and $h_v$ represents a node from a graph convolutional network (GCN). The value for such a graph can be determined by a recurrent neural network.
\begin{equation}
    h_{v}^{t} = f(W_1 h_{v}^{(t-1)} + W_2 \sum_{v^{'}\in N(v)}\alpha_{v^{'}}h_{v^{'}}^{(t-1)})
\end{equation}
for $t = 1,....,T$.

In a relational graph convolutional network (GCN), edges represent different relations of different types. Now, entities can be represented as $h_e^{(0)} = L(e) \in \mathbb{R}^p$ and documents can be represented as $h_d^{(0)} = LSTM(d_{w_1},...,d_{w_T}) \in \mathbb{R}^{T \times p}$ where $e$ stands for an entity and $d$ stands for a document. The propagation of information from entities to documents has been done using the equation as follows.
 
 \begin{equation}
     h_{d}^{(t)} = LSTM(h_{d_1}^{(t-1)}||e_{w_1}^{(t-1)},....,h_{d_T}^{(t-1)}||e_{wt}^{(t-1)})
 \end{equation}
 
 One the other way, the representation coming from the text can be passed into the entities.
 \begin{equation}
    \begin{aligned}
     h_{e}^{(t)} = f(W_1 h_{e}^{(t-1)}) + \sum_r \sum_{v^{'}\in N_{r}(v)} W_{2}^{r}h_{v^{'}}^{t-1} \\
     +W_3 \sum_{d:e \in d} h_{d_w}^{(t-1)}
     \end{aligned}
 \end{equation}

Authors have focused on the \textit{Early Fusion} strategy where a model is trained to extract answers from a \textit{question subgraph} containing relevant KB facts as well as text sentences. Graph representation learning has been used as a base framework. Modifications like \textit{heterogeneous update} and \textit{direct propagation} have been made on top of the basic graph-propagation model to make it adopted for the training. 

Domain knowledge can be integrated into the form of hierarchies which can be produced by stacked generalization. Such an integration on a logistic regression model can outperform the regular logistic regression model shown in \cite{b13}.

\section{Logical Domain Knowledge Integration}

In general, a machine learning model can be thought of as a statistical model, a generative model, or a classification model, which can be expressed as $x \sim p_\theta (x)$. Let's consider a constrain function in a neural network: $f_\Phi (x) \in  \mathbb{R}$, where $\Phi$ is a parameter in this network, which defines how much knowledge of $x$ has been embedded in the network. In common sense, a higher value of $\Phi$ indicates a better knowledge integration to the learning process.
Representation of logic in the vast space of neural networks exhibits different ideas of knowledge integration. Some of the key ideas researchers have worked on are described here.

\subsection{Structured Consistency as Constraint Function}
Incorporating structured domain knowledge into a deep generative model often requires complex modeling. Though posterior regularization (PR) provides a general framework to impose such structured constraints, it has limited applicability. Make relations between PR and reinforcement learning (RL) with the expansion of PR to learn constraints as rewards have been proposed in \cite{b12}.

Let's consider a generative model $ x \sim p_\theta(x) $ and a constraint function $f_\phi (x) \in \mathbb{R}$. A higher $f_\phi (x) \in \mathbb{R}$ value means a better $x$ corresponding to the knowledge. The main objective of imposing a constraint on a model is to make sure that the model is abide by the rules.
Integration of such a constraint can be done by increasing the expectation, $\mathbb{E}_{p\theta}[f(x)]$. What this equation expresses is to assign high probabilities to the input space where the constraints are met. Like in \cite{b12}, authors tried to generate images where the constraints meet mostly. Maximization of the objective can be done using the equation as follows,

\begin{equation}
    \underset{\theta}{min} (\mathcal{L(\theta)}- \alpha \mathbb{E}_{p\theta}[f_\phi(x)])
\end{equation}
where $\mathcal{L}$ denotes a regular loss function and the second term works as the regularization technique. But for a complex model $p$, the calculation of such a function can be difficult. One workaround is to find a variational distribution $q$, which is easy to compute and encouraged to stay close to the original $p$. Thus, the equation becomes:

\begin{equation}
    \mathcal{L}(\theta,q) = KL(q(x)||p_{\theta}(x) - \alpha \mathbb{E}_q[f(x)]) 
\end{equation}

The equation actually forces the $q$ distribution to be as close as $p$ using Kullback-Leibler Divergence(KL). $\alpha$ is the weight on the constraint term.

\begin{equation}
    \underset{\theta, q}{min} \mathcal{L}(\theta) + \lambda\mathcal(\theta, q)
\end{equation}
where $\lambda$ is the balancing hyper-parameter. 

The optimal solution of $q$ is as follows.
\begin{equation}
    q^{*}(x) = p_{\theta}(x)exp(\lambda f_{\theta}(x))/\mathcal{Z}
\end{equation}
where $Z$ is the normalization term.

The equation expresses the idea that a higher value of the constraint will result in a higher probability in $q$.

\subsection{Proposition Logic Encoding as A Constraint Function}

A statistical Relational Learning framework like Semantic Based Regularization (SBR) \cite{b8} can be used as an underlying framework for integrating prior knowledge into deep learning, which can be expressed as a collection of first-order logic (FOL). Experiments have been carried out on an image classification task where the aim is to determine the class of an animal. The optimized equation is like the following one.

\begin{equation}
    \begin{aligned}
    C_e [f(\mathcal{X})] = \sum_{k=1}{T}(||f_k||^2 + \lambda_l \sum_{x \epsilon \varepsilon_k}^{H} L(f_k(x), y_k(x)) 
    \\+\sum_{h=1}^{H}\lambda_h L_c(\Phi_h(f(\mathcal{X})))
    \end{aligned}
\end{equation} 
\newline
Where $\Phi_h(f), 0 \leq \Phi_h(f) \leq 1, h = 1, ...., H $ describes the prior knowledge. The first term of the equation is a regularization term, which actually restricts the solution. $\varepsilon_k$ is the set of labeled data. $L(.,.)$ is a loss function, $y_k(x)$ is the target output value for the $x$ pattern for task k, and $\lambda_l$ acts as the weight for the labeled portion of the cost function. $L_c (.)$ is the loss function that has been used for the constraint part, and $\lambda_h$ is the weight for the $h$-th constraint. 
First-order logic (FOL) rules can be expressed as propositional logic. To use propositional logic as a loss function, it has to be converted into a continuous and differentiable form. Basic logic operations $ (\neg, \lor, \land, \implies) $ of an expression tree representing a FOL can be replaced by unit computing the logic operation \cite{b8}. As an example, the output of the expression tree of the rule $\forall{x}[\neg A(x)] \lor [B(x) \land C(x)]$ can be expressed as $t_E(f(x)) = 1 - f_A(x) + f_A(x).f_B(x).f_C(x)$. 
The degree of the truth of a formula containing expression $E$ is the average of the norm generalization $t_{E}(.)$ where $x_i$ is the universally quantified variable. The knowledge function is defined as follows.

\begin{equation}
\begin{aligned}
    \Phi_h(f(\mathcal{X})) = \frac{1}{|\mathcal{X}_i|}\sum_{x_{i} \in \mathcal{X}_i} t_{E}(f([x, \mathcal{X}/\mathcal{X_i}]))
\end{aligned}
\end{equation}
where $\mathcal{X}_i$ is represented as the set of sample pattern inputs to the function.

Using the newly constructed grounded expression equation $t_E(f(x)) = 1 - f_A(x) + f_A(x).f_B(x).f_C(x)$, Equation (7) will become: 
\begin{equation}
\begin{aligned}
    \Phi_h(f(\mathcal{X})) = \sum_{x \in \mathcal{X}} t_E(f(x)) = \\
    = \sum_{x \in \mathcal{X}} 1 - f_A(x) + f_A(x).f_B(x).f_C(x)
\end{aligned}
\end{equation}

Embedding symbolic knowledge expressed as logic rules can be another form of knowledge injection into deep learning. In \cite{b18} the authors deployed a new technique of \textit{semantic regularization}, which can learn the embeddings semantically consistent with d-DNNF formulae as d-DNNF is more tractable than the CNF form. Two approaches have been followed in this experiment. The first one is the addition of logic loss as a means of applying soft constraints, and the other approach is to learn the vector-based representation of symbolic knowledge using neural networks.  
The layer-wise propagation rule of a Graph Convolution Network is defined as follows.

\begin{equation}
    Z^{(l+1)} = \sigma (\Tilde{D}^{- \frac{1}{2}} \Tilde{A}\Tilde{D}^{-\frac{1}{2}} Z^{(l)} W^{(l)})
\end{equation}
where $Z(l)$ are the latent node embeddings at $l^th$, $\Tilde{A} = A + I_N$ is the adjacency matrix of undirected graph, and $\Tilde{D}$ is a diagonal degree matrix.

Semantic regularization has been done over this structure by incorporating the d-DNNF logic graph that exhibits structural constraints. Children embeddings of $\land$ gates have been regularized to be orthogonal and $\lor$ gates to sum up to the unit vector. 
The resultant semantic regularization loss is as follows.

\begin{multline*}
    l_r (F) = \sum_{v_i \in \mathcal{N}_{o}}  \parallel { \sum\limits_{\substack{element-wise \\ v_j \in c_i }}} q(v_j) - 1 \parallel_{2}^{2} +
    \\  
    \sum_{v_k \in N_A} \parallel V_{k}^{T} - diag(V_{k}^{T}V_k) \parallel_{2}^{2}
\end{multline*}
\newline
where $q$ is our logic embedder, $\mathcal{N}_o$ is the set of $\lor$ nodes, $\mathcal{N}_A$ is the set of $\land$ nodes. $V_k = [q(v_1), q(v_2),.....,q(v_l)]$ where $v_1 \in C_k$.

To make it close to satisfying the assignments and far from unsatisfying assignments, the embedder has been trained with \textit{triplet loss}, which is defined as follows.

\begin{equation}
    \begin{aligned}
     l_t (F, \tau_T , \tau_F) \\= max\{d(q(F), q(\tau_F)) - d(q(F), q(\tau_T))+m,0 \} 
    \end{aligned}
\end{equation}
where $q(F)$ is the embedding of the d-DNNF logic graph for a given formula, $q(\tau_T)$ for satisfying, and $q(\tau_F)$ for unsatisfying as the assignments embeddings. Further training is done on model $h$ by augmenting the per-datum loss with a logic loss $l_logic$, which is defined as follows.

\begin{equation}
    l = l_c + \lambda l_{logic}
\end{equation}

\subsection{Semantic Loss Function as Constraint}
Outputs in neural networks can be connected through symbolic knowledge, where knowledge is represented as constraints in boolean logic. The semantic loss function identifies the closeness between the output of the neural network and its constraints. Significant improvement has been achieved in semi-supervised learning using this semantic loss function \cite{b14}. Semantic loss function can be expressed as $L^s (\alpha, p)$ where $\alpha$ is a sentence of proposition logic. The equation of a semantic loss function can be formulated as follows.
\begin{equation}
    L^{s} (\alpha, p) \propto - log \sum_{x \vDash \alpha} \underset{{i: x \vDash X_{i}}}{\invamalg}P_{i} \underset{i: x \vDash \neg X_{i}}{\invamalg}(1-P_{i})
\end{equation}
where $P$ is a probability vector for each $X$.
Thus, the new loss function will be
\begin{equation}
    existing loss + w.semantic loss
\end{equation}

\subsection{Approximation and Monotonicity as Constraint Function}
The performance of a neural network is questionable when a given data is either limited or noisy. Domain-based constraint integration as prior knowledge can surely amplify the performance in this regard. The representation of such knowledge has several forms. One form of such knowledge can be represented as the quantitative range of a normal operation, whereas monotonically increasing and decreasing relationships between different process variables or measurements of the same process but in different contexts can be another type of knowledge. The idea of imposing approximation and monotonicity as a constraint to the loss function to the network to improve the performance has been proposed in \cite{b9}.

Measurement can be noisy at times. Such kinds of measurements can cause significant deviations in the efficiency of a model. To make the model robust, insight domain knowledge can be helpful as it exhibits the reasonable ranges of a normal operation of the target variable. These inside-domain experts can be integrated into a model as approximation constraints. The functional form of such an approximation constraint can be defined as follows.

\begin{equation}
    g(\hat{Y}) = 
    \begin{cases}
    0 & \text{if $\hat{Y} \in [y_l, y_u]$} \\
    |y_l - \hat{Y}| & \text{if $\hat{Y}$ $<$ $ y_l$} \\
    |y_u - \hat{Y}| & \text{if $\hat{Y}$ $>$ $ y_u$}
    \end{cases}
\end{equation}

\begin{equation}
    Loss_D(\hat{Y}) = \sum^{m}_{i=1} ReLU(y_l - y_i) + ReLU(y_i - y_u)
\end{equation}
where $Y$ is the target variable,$(y_l, y_u)$ is the normal operation of a specific target variable, and $RELU$ is the short form for the rectified linear unit function. 

In real-world processes, like physical, chemical, and biological processes, monotonic relationships are presented between different entities. As an example, suppose $x_1$ and $x_2$ are two variables having a relation like $x_1 > x_2$. Now, consider a function $h(x) = y$. If for $x_1 > x_2$ the relationship between their corresponding function is $h(x_1) > h(x_2)$, then they are considered in a monotonic relation. We can find similar relations in real-world datasets. This idea can be implemented as a constraint to improve the efficiency of a machine learning model.   
A monotonicity constraint can be expressed as follows.
\begin{equation}
    \begin{aligned}
        loss_D(\hat{Y}_1, \hat{y}_2)=\\
        \sum_{i=1}^{m} ((x_{1}^{i} < x_{2}^{i}) \land (\hat{y}_{1}^{i} > \hat{y}_{2}^{i})).RELU(\hat{y}_{1}^{i} - \hat{y}_{2}^{i})
    \end{aligned}
\end{equation}

Integrating both equations into a normal loss function of a neural network has significantly improved the performance in \cite{b9}. Thus, the final loss function will be defined as follows.
\begin{equation}
    \arg\min_{f} Loss(Y, \hat{Y}) + \lambda_{D} Loss_{D}(\hat{Y}) + \lambda R(f)
\end{equation}
where $Loss(Y, \hat{Y})$ is the regular mean squared loss with $Y$ being the ground truth and $\hat{Y}$ being the predicted value.

\subsection{Logic Rules as Knowledge Distillation}

The uninterpretetability of neural network models can be reduced by combining logic rules into deep neural networks. It is a great way to consolidate the flexibility of models in a way to transfer human intention and domain knowledge to neural models. Such a framework has been proposed on \cite{b10} on a CNN experimenting sentiment analysis and an RNN for named entity recognition.
The conditional probability of a neural network can be defined as $p_\theta (\textbf{y}|\textbf{x})$. The $k$-dimensional soft prediction vector can be described as $\sigma_\theta (x)$. The network is parameterized by weight \textbf{${\theta}$} that means to produce the correct labels of training instances the network has to update \textbf{${\theta}$}. Now, to inject the domain knowledge or information presented as rules, the network is trained to mimic the outputs of a rule-regularized projection of $p_\theta (\textbf{y}|\textbf{x})$, which is denoted as $q_\theta (\textbf{y}|\textbf{x})$. To make a balance between imitating the soft predictions of $q$ and the prediction of true hard labels, the new formulas is established.

\begin{equation}
    \begin{aligned}
    \theta^{(t+1)} = arg min_{(\theta \in \Theta)} \frac{1}{N} \sum_{n=1}^{N} ({1- \pi}) l(y_n , \sigma_{\theta}(x_n))+\\ \pi l(s_n^{(t)}, \sigma_{\theta} (x_n))
    \end{aligned}
\end{equation}
where $l$ denotes the loss function in term of specific applications, $s_{n}^{(t)}$ is the vector of soft prediction $q$ on $x_n$ at iteration $t$, and $\pi$ denotes to calibrate relative importance between two objectives. Though the idea is motivated by the process called \textit{distillation}, the important difference has been made where the teacher and the student are learned simultaneously. Here $p_{\theta}(y|x)$ is the "student" and $q(y|x)$ is the "teacher". The construction of the teacher network is a vital point in this research. Two factors have been followed. First of all, for each rule and each of the groundings on $(X, Y)$, $\mathbb{E}_{q(\textbf{Y}|\textbf{X})}[r_{lg}(X, Y)] = 1$ with a confidence $\lambda_{l}$. For the second property, closeness has been measured between $q$ and $p_\theta$ with KL-divergence. Combining the two factors with the allowance of slackness for the constraints, the final optimization problem is formed as follows.
\begin{equation}
    \begin{aligned}
    min_(q, \xi \geq 0)KL(q(\textbf{Y}|\textbf{X})||p_{\theta}(\textbf{Y}||\textbf{})) + C \sum_{l, g_l} \xi_{l,g_l}\\
    s.t. \lambda_l(1 - \mathbb{E}[r_{l,g_l}(\textbf{X}, \textbf{Y})]) \leq \xi_{l,g_l}\\
    g_l = 1, ..., G_l, l = 1, ... , L,
    \end{aligned}
\end{equation}

\subsubsection{Knowledge Distillation for Resource Optimization}
The performance of a machine learning algorithm is measured in terms of its average performance on multiple models running over the same dataset. Instead of using multiple models, one way is to combine the models to build a robust model that can give an effective prediction. But the building, training, and testing of such a model are expensive and resource-heavy. Transferring some knowledge from a bigger model to a small and light model can give a significant performance enhancement \cite{b11}. 

In a different classification task, the probabilities of predicting a class are produced by the imposing "softmax" function on the output layer of the network. Softmax takes the logits and converts them to a probability by comparing them with other classes. 

\begin{equation}
    q_i = \frac{exp (z_i / T)}{\sum_j exp (z_i / T)}
\end{equation}\\
where $z_i$ is referred to as lots, and $T$ is the temperature variable to produce a softer probability distribution. 
The idea behind the teacher-student knowledge distillation architecture is to transfer knowledge from a complex model to a simple model. For example, transfer knowledge from the teacher model to the student model. The motivation behind the idea is to avoid making the one-hot encoded values as hard targets for the student model and to train it on a transfer set of softened probabilities that is predicted by the teacher model. 

\section{Scientific Knowledge Integration}

Scientific knowledge integration is another great aspect of incorporating domain knowledge into the neural network. 
For example, partial and stochastic differential equations like Newton's law of motion and Navier-Stokes fluid dynamics equations can be expressed in the form of domain knowledge. Different conversion laws are also part of scientific knowledge. 

\subsection{Law of Physics as Scientific Knowledge}
The Law of Physics can be embedded into a neural network as constraints. With the integration of such knowledge, a traditional convolutional neural network can be trained efficiently to predict objects without labeled data \cite{b17}.
The traditional neural function can be expressed as follows.

\begin{equation}
    f^* = \underset{f \in F}{argmin} \sum_{i=1}^{n} l(f(x_i), y_i))
\end{equation}

The authors \cite{b17} have carried out two experiments in this regard. The motivation behind the first experiment is to learn an object's height from the videos where the object has been thrown across the field of view. The equation that has been counted here as a constrain function is as follows.

\begin{equation}
    y_i = y_o + v_0(i \Delta t) + a(i \Delta t)^2
\end{equation}\\
where $\Delta t$ = 0.1s denotes the duration between frames. 
This function is now integrated as a loss function into the network. The combined equation is represented as follows.
\begin{equation}
    g(x, f(x)) = g(f(x)) = \sum_{i=1}^{N} |\hat{Y}_i - f(x_i)|
\end{equation}

In the other experiments, the authors have done experiments on tracking the position of a walking man where the velocity is considered as a constant over a period of time. 

\subsection{Space Knowledge as Scientific Knowledge}

Deep  learning  has  already  made  its  mark on Space-based missions. Using the state-of-the-art deep learning models, Shallue and Vanderburg has been successful in  automatically classifying Kepler transit signals as exoplanets or false positives \cite{b15}.  The idea of integrating scientific domain knowledge has taken this work to the next level with the significant improvement of overall performance in \cite{b16}. On top of that, data augmentation techniques have been developed to alleviate the model over-fitting, which  has helped reduce the size of the model, maintaining its performance. 

\section{Discussion}
The unavailability of vast amounts of data is the main hindrance in the field of machine learning. Poor quality of data, cost of data, and lacking data all lead to the poor performance of a machine learning model. Integrating domain knowledge into machine learning models can come out with great success in this regard. However, challenges arise while incorporating such knowledge into a neural network. Here, we will discuss the drawbacks of some of the existing works and how things can be improved to provide a direction for future research.

Embedding symbolic knowledge into a graph convolutional neural network is an amazing concept. Expressing logical formula into a Deterministic Decomposable Negation Normal Form (d-DNNF) formulae \cite{b26} and embedding into a GCN, a significant performance boost has been achieved in \cite{b18}. But apart from d-DNNF, there are other forms of symbolic knowledge that keep the door open for further improvement. Also, for relatively scarce data, the performance is still questionable. 

We saw that logical rules like propositional logic can be a great source of knowledge that can be incorporated into machine learning modules. But the experiment has only been done using certain models. It is still unsure how the performance will vary using other deep architectures. On top of that, different problems need different propositional logic embedded into them. The use of an enormous amount of propositional logic in a neural network can affect performance. Is it justified to trade off the performance is still questionable. 

Scientific knowledge addition to deep learning is another outstanding idea that has been applied by the NASA research team to automatically classify \textit{Kepler} candidate transit events. On top of that, the inclusion of expert domain knowledge in state-of-the-art deep learning models can be used to identify weak signals in noisy data. Space-based photometry missions that focus on finding small planets, such as \textit{TESS} and \textit{PLATO}, can get significant help from the idea of domain knowledge integration.

Tasks like open-domain question answering have been popular in recent times in the field of natural language processing. We know that only using unstructured text cannot be fruitful sometimes as this technique uses shallow methods. Using only a knowledge graph is also not practical because of noisy data and missing information. Combining both processes, following the "early fusion" technique the performance can be much improved. Such work has been done by the GRAFT-Net framework in \cite{b7}. But the sub-graph retrieval process still remains too complicated at times. A significant improvement can be done with early fusion to make the model easier in integrating knowledge.

\section{Conclusion}
In conclusion, the main limitation of knowledge integration is that knowledge has been represented so far in the forms of relational, logical, and simple constraints. But human knowledge is more abstract, robust, and built on vast experience and high-level concepts. There are no denying facts that knowledge integration won't improve the performance of a model. The main difficult part is how to encode such knowledge and how efficiently it can be done. The key features still rely on the fact that how relevant these constraints can be from task to task. Because it won't make any impact if the constraints are irrelevant to the corresponding dataset.

It still remains an open question about whether we actually need domain knowledge or just pick a simple model and scale it up to get the desired output. But if we really dig deeply into the proposed techniques in incorporating knowledge in machine learning, we can easily figure out that domain knowledge can indeed improve the performance of a model regardless of the scale of the machine learning models.   

\bibliographystyle{IEEEtran}
\bibliography{references}

\end{document}